\documentclass[11pt,a4paper]{article}
\usepackage[hyperref]{eacl2021}
\usepackage{times} 
\usepackage{helvet} 
\usepackage{courier} 
\usepackage{url} 
\usepackage{graphicx} 
\usepackage{natbib}
\usepackage{caption}
\usepackage{times}
\usepackage{latexsym}

\usepackage{microtype}
\usepackage{esvect}
\usepackage{latexsym}
\usepackage{mathtools}
\usepackage{tcolorbox}
\usepackage{breqn}
\usepackage{graphicx}
\usepackage{algorithm}
\usepackage{algorithmic}
\usepackage{esvect}
\usepackage{latexsym}
\usepackage{mathtools}
\usepackage{tcolorbox}
\usepackage{breqn}
\usepackage{makecell}

\usepackage{multirow}
\usepackage{amsmath}
\usepackage{amssymb}
\usepackage{subcaption}
\usepackage{booktabs}
\usepackage{array}
\usepackage{threeparttable}

\usepackage{array}
\usepackage{placeins}
\newcolumntype{?}{!{\vrule width 1pt}}

\aclfinalcopy 
\def\aclpaperid{803} 

\title{Multilingual and cross-lingual document classification:\\A meta-learning approach}
  
 \author{Niels van der Heijden$^{\clubsuit}$ ~  Helen Yannakoudakis$^{\spadesuit}$ ~Pushkar Mishra$^{\diamondsuit}$ ~ Ekaterina Shutova$^{\clubsuit}$\\
$^\clubsuit$ILLC, University of Amsterdam, the Netherlands \\
$^\spadesuit$Dept. of Informatics, King's College London, United Kingdom \\
$^\diamondsuit$Facebook AI, London, United Kingdom \\

{ \small \tt niels.vanderheijden@student.uva.nl, helen.yannakoudakis@kcl.ac.uk, } \\
{ \small \tt pushkarmishra@fb.com, e.shutova@uva.nl}
}

\begin{document}
\maketitle
\begin{abstract}
   The great majority of languages in the world are considered under-resourced for the successful application of deep learning methods. In this work, we 
   propose a meta-learning approach to document classification in a limited-resource setting and demonstrate its effectiveness in two different settings: few-shot, cross-lingual adaptation to previously unseen languages; and multilingual joint training when limited target-language data is available during training. 
   We conduct a systematic comparison of several   
   meta-learning methods, investigate multiple settings in terms of data availability and show that meta-learning thrives in settings with a heterogeneous task distribution. We propose a simple, yet effective adjustment to existing meta-learning methods which allows for better and more stable learning, and set a new state of the art on several languages while performing on-par on others, using only a small amount of labeled data.
\end{abstract}

\section{Introduction}
There are more than 7000 languages around the world and, of them, around 6\% account for 94\% of the population.\footnote{https://www.ethnologue.com/statistics}
Even for the 6\% most spoken languages, very few of them possess adequate resources for natural language research and, when they do, resources in different domains are highly imbalanced. Additionally, human language is dynamic in nature: new words and domains emerge continuously and hence no model learned in a particular time will remain valid forever. 

With the aim of extending the global reach of Natural Language Processing (NLP) technology, much recent research has focused on the development of multilingual models and methods to efficiently transfer knowledge across languages. Among these advances are multilingual word vectors which aim to give word-translation pairs a similar encoding in some embedding space \cite{mikolov2013exploiting,lample2017unsupervised}. There has also been a lot of work on multilingual sentence and word encoders that either explicitly utilizes corpora of bi-texts \cite{artetxe2019massively,lample2019cross} or jointly trains language models for many languages in one encoder \cite{devlin2018bert,conneau2019unsupervised}. Although great progress has been made in cross-lingual transfer learning, these methods either do not close the gap with performance in a single high-resource language \cite{artetxe2019massively,conneau2019unsupervised,van2019comparison}, e.g., because of cultural differences in languages which are not accounted for, or are impractically expensive \cite{lai2019bridging}. 

Meta-learning, or \textit{learning to learn} \cite{schmidhuber1987evolutionary,bengio1990learning,thrun1998learning}, is a learning paradigm which focuses on the quick adaption of a learner to new tasks. The idea is that by training a learner to adapt quickly and from a few examples on a diverse set of training tasks, the learner can also generalize to unseen tasks at test time. Meta-learning has recently emerged as a promising technique for few-shot learning for a wide array of tasks \cite{finn2017model,koch2015siamese,ravi2016optimization} including NLP \cite{dou2019investigating,gu2018meta}.  
To our best knowledge, no previous work has been done in investigating meta-learning as a framework for multilingual and cross-lingual few-shot learning. We propose such a framework and demonstrate its effectiveness in document classification tasks. The only current study on meta-learning for cross-lingual few-shot learning is the one by \cite{nooralahzadeh2020zero}, focusing on natural language inference and multilingual question answering. In their work, the authors focus on applying meta-learning to learn to adapt a monolingually trained classifier to new languages. In contrast to this work, we instead show that, in many cases, it is more favourable to not initialize the meta-learning process from a monolingually trained classifier, but rather reserve its respective training data for meta-learning instead.

Our contributions are as follows: 1) We propose a meta-learning approach to few-shot cross-lingual and multilingual adaptation and demonstrate its effectiveness on document classification tasks over traditional supervised learning; 
2) We provide an extensive comparison of meta-learning methods on multilingual and cross-lingual few-shot learning 
    and release our code to facilitate further research in the field;\footnote{\tiny{\url{https://github.com/mrvoh/meta_learning_multilingual_doc_classification}}} 
3) We analyse the effectiveness of meta-learning under a number of different parameter initializations and multiple settings in terms of data availability, and show that meta-learning can effectively learn from few examples and diverse data distributions;
    4) We introduce a simple yet effective modification to existing methods and empirically show that it stabilizes training and converges faster to better local optima; 
    5) We set a new state of the art on several languages and achieve on-par results on others using only a small amount of data. 

\section{Meta-learning methods}
\begin{algorithm}[t]
{\footnotesize
\caption{\label{alg:main} Meta-training procedure.}
\begin{algorithmic}
    \REQUIRE{$p(\mathcal{D})$: distribution over tasks.}
    \REQUIRE{$\alpha, \beta$: step size hyper-parameters}
    \STATE{Initialize $\theta$}
    \WHILE{not done}
        \STATE{Sample batch of tasks $\{D^l\} = \{(S^l, Q^l)\} \sim p(\mathcal{D})$}
        \FORALL{$(S^l, Q^l)$}
            \STATE{Initialize $\theta_l^{(0)} = \theta$}
            \FORALL{steps k}
                \STATE{
                \begin{flushleft}Compute: $\theta_l^{(k+1)} =\theta_l^{(k)} - \alpha( \nabla_{\theta_l^{(k)}} \mathcal{L}_{S_l}(f_{\theta_l^{(k)}}))$\end{flushleft}}
            \ENDFOR
        \ENDFOR
        \STATE{Update $\theta = \theta - \beta($MetaUpdate$(f_{\theta_l^{(K)}}, Q^l))$} \label{algo:meta-update}
	\ENDWHILE
\end{algorithmic}
}
\end{algorithm}
\noindent Meta-learning, or \textit{learning to learn}, aims to create models that can learn new skills or adapt to new tasks rapidly from few training examples. Unlike traditional machine learning, datasets for either training or testing, which are referred to as \textit{meta-train} and \textit{meta-test} datasets, comprise of many tasks sampled from a distribution of tasks $p(\mathcal{D})$ rather than individual data points. Each task is associated with a dataset $\mathcal{D}$ which contains both feature vectors and ground truth labels and is split into a \textit{support set} and a \textit{query set}, $\mathcal{D} = \{S,Q\}$. The support set is used for fast adaptation and the query set is used to evaluate performance and compute a loss with respect to model parameter initialization. 
Generally, some model $f_{\theta}$ parameterized by $\theta$, often referred to as the \textit{base-learner}, is considered. 
 A cycle of fast-adaptation on a support-set followed by updating the parameter initialization of the base-learner based on the loss on the query-set is called an \textit{episode}. 
In the case of classification, the optimal parameters maximize the probability of the true labels across multiple batches $Q \subset \mathcal{D}$ 
\begin{align}\displaystyle
    \theta^* := arg\underset{\theta}{max} \mathbb{E}_{Q \subset \mathcal{D}}[\sum_{(x,y) \in Q}P_\theta(y|x)]
\end{align}
In few-shot classification/fast learning, the goal is to minimize the prediction error on data samples with unknown labels given a small support set for learning. 
\textit{Meta-training} (Algorithm \ref{alg:main}) consists of updating the parameters of the base-learner by performing many of the formerly described \textit{episodes}, until some stop criterion is reached.

Following this procedure, the extended definition of optimal parameters is given in Eq.~\ref{eq:extended_def_optimal_params} to include fast adaptation based on the support set. The underlined parts mark the difference between traditional supervised-learning and meta-learning. The optimal parameters $\theta^* $ are obtained by solving
\begin{align}\scriptstyle
    arg\underset{\theta}{max} \underline{\mathbb{E}_{l \subset L}[}\mathbb{E}_{\underline{S^l \subset \mathcal{D}}, Q^l \subset \mathcal{D}}[\sum_{(x,y) \in Q^l}P_\theta(y|x, \underline{S^l})]\underline{]} \label{eq:extended_def_optimal_params}
\end{align}
In this work, we focus on metric- and optimization-based meta-learning algorithms. In the following sections, their respective characteristics and the update methods in Algorithm \ref{alg:main} are introduced. 
\subsection{Prototypical Networks}
Prototypical Networks \cite{snell2017prototypical} belong to the metric-based family of meta-learning algorithms. Typically they consist of an embedding network $f_\theta$ and a distance function $d(x_1, x_2)$ such as Euclidean distance. The embedding network is used to encode all samples in the support set $S_c$ and compute \textit{prototypes} $\mu_c$ per class $c \in C$ by computing the mean of the sample encodings of that respective class
\begin{align}
    \mu_c := \frac{1}{|S_c|}\sum_{(x_i, y_i) \in S_c} f_\theta(x_i) \label{eq:prototypes}
\end{align}

Using the computed prototypes, Prototypical Networks classify a new sample as
\begin{align}
    p(y=c|x) = \frac{exp(-d(f_\theta(x), \mu_c)}{\sum_{c^' \in C} exp(-d(f_\theta(x), \mu_{c^'})}
\end{align}
\citet{wang2019simpleshot} show that despite their simplicity, Prototypical Networks can perform on par or better than other state-of-the-art meta-learning methods when all sample encodings are centered around the overall mean of all classes and consecutively L2-normalized. We also adopt this strategy. 

\subsection{MAML}
Model-Agnostic Meta-Learning (MAML) \cite{finn2017model} is an optimization-based method that uses the following objective function
\begin{align}
    \theta^* := arg\underset{\theta}{min}\sum_{D_l \sim p(D)}\mathcal{L}_l(f_{\theta_l^{(k)}})
\end{align}
$\mathcal{L}_l(f_{\theta_l^{(k)}})$ is the loss on the query set after updating the base-learner for $k$ steps on the support set. Hence, MAML directly optimizes the base-learner such that fast-adaptation of $\theta$, often referred to as \textit{inner-loop optimization}, results in task-specific parameters $\theta_l^{(k)}$ which generalize well on the task.
Setting $B$ as the batch size, MAML implements its MetaUpdate, which is also referred to as \textit{outer-loop optimization}, 
as
\begin{align}
    \theta = \theta - \beta \frac{1}{B}\sum_{D_l \sim p(\mathcal{D})}(\nabla_\theta \mathcal{L}_l(f_{\theta_l^{(k)}}))
\end{align}

\noindent Such a MetaUpdate requires computing second order derivatives and, in turn, holding $\theta_l^{(j)} \forall j = 1, \dots , k$ in memory. A first-order approximation of MAML (foMAML), which ignores second order derivatives, can be used to bypass this problem:
\begin{align}
    \theta = \theta - \beta \frac{1}{B}\sum_{D_l \sim p(\mathcal{D})}(\nabla_{\theta_l^{(k)}} \mathcal{L}_l(f_{\theta_l^{(k)}}))
\end{align}
Following previous work \cite{antoniou2018train}, we also 
adopt the following improvements in our framework for all MAML-based methods: 
     \paragraph{Per-step Layer Normalization weights} Layer normalization weights and biases are not updated in the inner-loop. Sharing one set of weights and biases across inner-loop steps implicitly assumes that the feature distribution between layers stays the same at every step of the inner optimization. 
    \paragraph{Per-layer per-step learnable inner-loop learning rate} Instead of using a shared learning rate for all parameters, the authors propose to initialize a learning rate per layer and per step and jointly learn their values in the MetaUpdate steps.
    \paragraph{Cosine annealing of outer-loop learning rate} It has shown to be crucial to model performance to anneal the learning rate using some annealing function  \cite{loshchilov2016sgdr}.

\subsection{Reptile}
Reptile \cite{nichol2018first} is a first-order optimization-based meta-learning algorithm which is designed to move the weights towards a manifold of the weighted averages of task-specific parameters $\theta_l^{(k)}$:
\begin{align}
    \theta = \theta - \beta\frac{1}{B}\sum_{D^l \sim p(\mathcal{D})}(\theta_l^{(k)} - \theta)
\end{align}
Despite its simplicity, it has shown competitive or superior performance against MAML, e.g., on Natural Language Understanding \cite{dou2019investigating}.

\subsection{ProtoMAML}
\citet{triantafillou2020meta} introduce ProtoMAML as a meta-learning method which combines the complementary strengths of Prototypical Networks and MAML by leveraging the inductive bias of the use of prototypes instead of random initialization of the final linear layer of the network. 
\citet{snell2017prototypical} show that Prototypical Networks are equivalent to a linear model when Euclidean distance is used. Using the definition of prototypes $\mu_c$ as per  Eq. \ref{eq:prototypes}, the weights $w_c$ and bias $b_c$ corresponding to class $c$ can be computed as follows
\begin{align}
    \mathbf{w}_c := 2 \mu_c \qquad b_c := - \mu_c^T\mu_c \label{eq:protomaml}
\end{align}

ProtoMAML is defined as the adaptation of MAML where the final linear layer is parameterized as per  
Eq. \ref{eq:protomaml} at the start of each episode using the 
support set. Due to this initialization, it 
allows modeling a varying number of classes per episode. 

\paragraph{ProtoMAMLn}
Inspired by \citet{wang2019simpleshot}, we propose a simple, yet effective adaptation to ProtoMAML by applying $L_2$ normalization to the prototypes themselves, referred to as ProtoMAMLn, and, again, use a first-order approximation (foProtoMAMLn). We demonstrate that doing so leads to a more stable, faster and effective learning algorithm at only constant extra computational cost ($\mathcal{O}(1))$. 

We hypothesize the normalization to be particularly beneficial in case of a relatively high-dimensional final feature space -- in case of BERT-like models typically 768 dimensions. Let $x$ be a sample and $\hat{x} = f_\theta(x)$ be the encoding of the sample in the final feature space. Since the final activation function is the \textit{tanh} activation, all entries of both $\hat{x}$ and $\mu_c$ have values between -1 and 1. The pre-softmax activation for class $c$ is computed as $\hat{x}^T\mu_c$. Due to the size of the vectors and the scale of their respective entries, this in-product can yield a wide range of values, which in turn results in relatively high loss values, making the inner-loop optimization unstable.

\section{Related work} \label{sec:rel work}
\subsection{Multilingual NLP}
Just as the deep learning era for monolingual NLP started with the invention of dense, low-dimensional vector representations for words \cite{mikolov2013distributed} so did cross-lingual NLP with works like those of \citet{mikolov2013exploiting,faruqui2014retrofitting}. More recently, multilingual and/or cross-lingual NLP is approached by training one shared encoder for multiple languages at once, either by explicitly aligning representations with the use of parallel corpora \cite{artetxe2019massively,lample2019cross} or by jointly training on some monolingual language model objective, such as the Masked Language Model (MLM) \cite{devlin2018bert}, in multiple languages \cite{devlin2018bert,conneau2019unsupervised}. 

The formerly described language models aim to create a shared embedding space for multiple languages with the hope that fine-tuning in one language does not degrade performance in others. \citet{lai2019bridging} argue that just aligning languages is not sufficient to generalize performance to new languages due to the phenomenon they describe as \textit{domain drift}. Domain drift accounts for all differences for the same tasks in different languages which cannot be captured by a perfect translation system, such as differences in culture. They instead propose a multi-step approach which utilizes a multilingual teacher trained with Unsupervised Data Augmentation (UDA) \cite{xie2019unsupervised} to create labels for a student model that is pretrained on large amounts of unlabeled data in the target language and domain using the MLM objective. With their method, the authors obtain state-of-the-art results on the MLDoc document classification task \cite{SCHWENK18.658} and the Amazon Sentiment Polarity Review task \cite{prettenhofer-stein-2010-cross}.
A downside, however, is the high computational cost involved. For every language and domain combination: 1) a machine translation system has to be inferred on a large amount of unlabeled samples; 2) the UDA method needs to be applied to obtain a teacher model to generate pseudo-labels on the unlabeled in-domain data; 3) a language model must be finetuned, which involves forwards and backwards computation of a softmax function over a large output space (e.g., 50k tokens for mBERT and 250k tokens for XLM-RoBERTa). The final classifier is then obtained by 4) training the finetuned language model on the pseudo-labels generated by the teacher.

\subsection{Meta-learning in NLP}
\paragraph{Monolingual}
\citet{bansal2019learning} apply meta-learning to a wide range of NLP tasks within a monolingual setting and show superior performance for parameter initialization over self-supervised pretraining and multi-task learning. Their method is an adaptation of MAML where a combination of a text-encoder, BERT \cite{devlin2018bert}, is coupled with a parameter generator that learns to generate task-dependent initializations of the classification head such that meta-learning can be performed across tasks with disjoint label spaces. \citet{obamuyide-vlachos-2019-model} apply meta-learning on the task of relation extraction; \citet{obamuyide-vlachos-2019-meta} apply lifelong meta-learning for relation extraction; \citet{chen-etal-2019-meta} apply meta-learning for few-shot learning on missing link prediction in knowledge graphs.
\paragraph{Multilingual}
\citet{gu2018meta} apply meta-learning to  Neural Machine Translation (NMT) and show its advantage over strong baselines such as cross-lingual transfer learning. By viewing each language pair as a task, the authors apply MAML to obtain competitive NMT systems with as little as 600 parallel sentences. 
To our best knowledge, the only application of meta-learning for cross-lingual few-shot learning is the one by \citet{nooralahzadeh2020zero}. The authors study the application of X-MAML, a MAML-based variant, to cross-lingual Natural Language Inference (XNLI) \cite{conneau2018xnli} and Multilingual Question Answering (MLQA) \cite{lewis2019mlqa} in both a cross-domain and cross-language setting. 
X-MAML works by pretraining some model $M$ on a high-resource task $h$ to obtain initial model parameters $\theta_{mono}$. Consecutively, a set $L$ of one or more auxiliary languages is taken, and MAML is applied to achieve fast adaptation of $\theta_{mono}$ for $l \in L$. In their experiments, the authors use either one or two auxiliary languages and evaluate their method in both a zero- and few-shot setting. It should be noted that, in the few-shot setting, the full development set (2.5k instances) is used to finetune the model,
which is not in line with other work on few-shot learning, such as \cite{bansal2019learning}. Also, there is a discrepancy in the training set used for the baselines and their proposed method. All reported baselines are either zero-shot evaluations of $\theta_{mono}$ or of $\theta_{mono}$ finetuned on the development set of the target language, whereas their proposed method additionally uses the development set in either one or two auxiliary languages during meta-training.

\begin{table*}[t]
\footnotesize
\centering
\begin{tabular}{llllll}
\toprule
\textbf{MetaUpdate Method} & \textbf{Num inner-loop steps}  & \textbf{Inner-loop lr} & \textbf{Class-head lr multiplier} & \textbf{Inner-optimizer lr} \\ 
                       \midrule
Reptile   & 2,3,\underline{5} & 1e-5, \underline{5e-5}, 1e-4 & \underline{1}, 10 & -  \\
foMAML & 2,3,\underline{5} & \underline{1e-5}, 1e-4, 1e-3  &  1, \underline{10} & 3e-5, \underline{6e-5}, 1e-4 \\
foProtoMAMLn & 2,3,\underline{5} & \underline{1e-5}, 1e-4, 1e-3  &  1, \underline{10} & 3e-5, \underline{6e-5}, 1e-4 \\
\bottomrule                       
\end{tabular}
\caption{Search range per hyper-parameter. We consider the number of update steps in the inner-loop, \textit{Num inner-loop steps}, the (initial) learning rate of the inner-loop, \textit{Inner-loop lr}, the factor by which the learning rate of the classification head is multiplied, \textit{Class-head lr multiplier}, and, if applicable, the learning rate with which the inner-loop optimizer is updated, \textit{Inner-optimizer lr}. The chosen value is underlined.}
\label{tab:grid_search}
\end{table*}

\section{Data} \label{sec:data}
In this section, we give an overview of the datasets we use and the respective classification tasks. 

\paragraph{MLDoc}
\citet{SCHWENK18.658} published an improved version of the Reuters Corpus Volume 2 \cite{lewis2004rcv1} with balanced class priors for all languages.  
MLDoc consists of news stories in 8 languages: 
English, Spanish, French, Italian, Russian, Japanese and Chinese. 
Each news story is manually classified into one of four groups: \textit{Corporate/Industrial, Economics, Government/Social} and \textit{Markets}. The train datasets contain 10k samples whereas the test sets contain 4k samples. 

\paragraph{Amazon Sentiment Polarity} \label{sec:amazon_data}
Another widely used dataset for cross-lingual text classification is the Amazon Sentiment Analysis dataset \cite{prettenhofer-stein-2010-cross}. The dataset is a collection of product reviews in English, French, German and Japanese in three categories: \textit{books} \textit{dvds} and \textit{music}. Each sample consists of the original review accompanied by meta-data such as the rating of the reviewed product expressed as an integer on a scale from one to five. In this work, we consider the sentiment polarity task where we distinguish between positive (rating $>$ 3) and negative (rating $<$ 3) reviews. 
When all product categories are concatenated, the dataset consists of 6K samples per language per dataset (train, test). We extend this with Chinese product reviews in the cosmetics domain from JD.com \cite{zhang2015daily}, a large e-commerce website in China. The train and test sets contain 2k and 20k samples respectively.

\section{Experiments}
We use XLM-RoBERTa \cite{conneau2019unsupervised}, a strong multilingual model, as the base-learner in all models. 
We quantify the strengths and weaknesses of meta-learning as opposed to traditional supervised learning in both a cross- and a multilingual joint-training setting with limited resources.
\paragraph{Cross-lingual adaptation}
Here, the available data is split into multiple subsets: the auxiliary languages $l_{aux}$ which are used in meta-training, the validation language $l_{dev}$ which is used to monitor performance, and the target languages $l_{tgt}$ which are kept \textit{unseen} until meta-testing. Two scenarios in terms of amounts of available data are considered. A small sample of the available training data of $l_{aux}$ is taken to create a limited-resource setting, whereas all available training data of $l_{aux}$ is used in a high-resource setting. The chosen training data per language is split evenly and stratified over two disjoint sets from which the meta-training support and query samples are sampled, respectively. For meta-testing, one batch (16 samples) is taken from the training data of each target language as support set, while we test on the whole test set per target language (i.e., the query set).

\paragraph{Multilingual joint training}
We also investigate meta-learning as an approach to multilingual joint-training in the same limited-resource setting as previously described for the cross-lingual experiments. The difference is that instead of learning to generalize to $l_{tgt} \ne l_{aux}$ from few examples, here $l_{tgt} = l_{aux}$. If we can show that one can learn many similar tasks across languages from few examples per language, using a total number of examples in the same order of magnitude as in ``traditional'' supervised learning for training a monolingual classifier, this might be an incentive to change data collection processes in practice. 

For both experimental settings above, we examine the influence of additionally using all training data from a high-resource language $l_{src}$ during meta-training, English. 

\begin{table*}[ht]
\centering
\resizebox{\textwidth}{!}{\begin{tabular}{llllllll|l|llllll|l}
\toprule
\multicolumn{1}{l}{\multirow{2}{*}{$\mathbf{l_{src}}$ = {en}}} & \multicolumn{1}{c}{\multirow{2}{*}{\textbf{Method}}} & \multicolumn{6}{c}{\textbf{Limited-resource setting}} & \multicolumn{6}{c}{\textbf{High-resource setting}}  \\
 & & \textbf{de}  & \textbf{fr} & \textbf{it} & \textbf{ja} & \textbf{ru} & \textbf{zh} & $\Delta$ & \textbf{de}  & \textbf{fr} & \textbf{it} & \textbf{ja} & \textbf{ru} & \textbf{zh} & $\Delta$ \\
                      \midrule
\multirow{5}{*}{Excluded} 
& Non-episodic & 82.0 & \textbf{86.7} & 68.3 & 71.9 & 70.9 & 81.0 & 76.8 & 95.3 & 90.9 & 80.9 & 82.9 & 74.5 & \textbf{89.6} & 85.7 \\
& ProtoNet  & 90.5 & 85.0 & 76.6 & 75.0 & 69.6 & 82.0 & 79.8 & 95.5 & 91.7 & 82.0 & 82.2 & 76.6 & 87.4 & 85.9 \\
& foMAML           & 89.7 & 85.5 & 74.1 & 74.1 & \textbf{74.0} & 83.2 & 80.1  & 95.0 & 91.4 & 81.4 & 82.7 & 76.9 & 87.8 & 86.1  \\
& foProtoMAMLn        & \textbf{90.6} & 86.2 & \textbf{77.8} & \textbf{75.6} & 73.6 & \textbf{83.8} & \textbf{80.7}  & \textbf{95.6} & \textbf{92.1} & \textbf{82.6} & \textbf{83.1} & \textbf{77.9} & 88.9 & \textbf{86.7} \\
& Reptile       & 87.9 & 81.8 & 72.7 & 74.4 & 73.9 & 80.9 & 78.6 & 95.0 & 90.1 & 81.1 & 82.7 & 72.5 & 88.7 & 85.0  \\ 
\midrule
\multirow{5}{*}{Included} 
& Zero-shot & 92.4 & \textbf{92.1} & 80.3 & 81.0 & 71.7 & 89.1 & 84.4 & 92.4 & 92.1 & 80.3 & 81.0 & 71.7 & 89.1 & 84.4 \\
& Non-episodic  & 93.7 & 91.3 & \textbf{81.5} & 80.6 & 71.1 & 88.4 & 84.4 & 93.7 & 92.9 & 82.4 & 82.3 & 72.1 & 90.1 & 85.6 \\
& ProtoNet & 93.4 & 91.9 & 79.1 & 81.3 & 72.2 & 87.8 & 84.5 & 95.0 & 91.7 & 81.1 & 82.7 & 72.0 & 88.0 & 85.9 \\
& foMAML           & \textbf{95.1} & 91.2 & 79.5 & 79.6 & 73.3 & 89.7 & 84.6  & 94.8 & 93.2 & 79.9 & 82.4 & 75.7 & \textbf{90.6} & 86.1  \\
& foProtoMAMLn   & 94.9 & 91.7 & \textbf{81.5} & \textbf{81.4} & \textbf{75.2} & \textbf{89.9} & \textbf{85.5}  & \textbf{95.8} & \textbf{94.1} & \textbf{82.7} & \textbf{83.0} & \textbf{81.2} & 90.4 & \textbf{87.9} \\
& Reptile       & 92.3 & 91.4 & 79.7 & 79.5 & 71.8 & 88.1 & 83.8 & 94.8 & 91.0 & 80.2 & 82.0 & 72.7 & 89.9 & 85.1 \\ 
\bottomrule                       
\end{tabular}}
\caption{{Average accuracy of 5 different seeds on the unseen target languages for MLDoc.}  $\Delta$ corresponds to the average accuracy across test languages. } 
\label{tab:results_mldoc_half}
\end{table*}
\subsection{Specifics per dataset}
\paragraph{MLDoc}
As MLDoc has sufficient languages, we set $l_{src} =$ English and $l_{dev} =$ Spanish. The remaining languages are split in two groups: $l_{aux}=\{\textrm{German, Italian, Japanese}\}$; and $l_{tgt}=\{\textrm{French, Russian, Chinese}\}$.
In the limited-resource setting, we randomly sample 64 samples per language in $l_{aux}$ for training. Apart from comparing low- and high-resource settings, we also quantify the influence of augmenting the training set $l_{aux}$ with a high-resource source language $l_{src}$, English.
\paragraph{Amazon Sentiment Polarity}
The fact that the Amazon dataset (augmented with Chinese) comprises of only five languages has some implications for our experimental design. 
In the cross-lingual experiments, where $l_{aux}$, $l_{dev}$ and $l_{tgt}$ should be disjoint, only three languages, including English, remain for meta-training. As we consider two languages too little data for meta-training, we do not experiment with leaving out the English data.
 Hence, for meta-training, the data consists of $l_{src} =$ English, as well as two languages in $l_{aux}$. We always keep one language unseen until meta-testing, and alter $l_{aux}$ such that we can meta-test on every language. 
 We set $l_{dev} =$ French in all cases except when French is used as the target language; then, $l_{dev} =$ Chinese. 
 In the limited-resource setting, a total of 128 samples per language in $l_{aux}$ is used. 
 
 For the multilingual joint-training experiments there are enough languages available to quantify the influence of English during meta-training. When English is excluded, it is used for meta-validation. When included, we average results over two sets of experiments: one where $l_{dev} =$ French and one where $l_{dev} =$ Chinese.
 
\begin{table*}[th]
\centering
\begin{tabular}{lllll|l|llll|l}
\toprule
 \multicolumn{1}{c}{\multirow{2}{*}{\textbf{Method}}} & \multicolumn{4}{c}{\textbf{Limited-resource setting}} & & \multicolumn{4}{c}{\textbf{High-resource setting}}  \\ & \textbf{de} & \textbf{fr} & \textbf{ja} & \textbf{zh} & $\Delta$  & \textbf{de} & \textbf{fr} & \textbf{ja} & \textbf{zh} & $\Delta$ \\
                       \midrule
Zero-shot & \textbf{91.2} & 90.7 & 87.0 & 84.6 & 88.4 & 91.2 & 90.7 & 87.0 & 84.6 & 88.4 \\
 Non-episodic & 90.9 & 90.6 & 86.1 & 86.9 & 88.6 & 91.6 & 91.0 & 85.5 & 87.9 & 89.0 \\
ProtoNet & 89.7 & 90.2 & 86.6 & 85.2 & 87.9 & 90.7 & 92.0 & 86.7 & 84.0 & 88.4\\
foMAML & 88.3 & 90.5 & 86.8 & 88.1 & 88.4 & 91.4 & 92.5 & 88.0 & \textbf{90.4} & 90.6 \\
foProtoMAMLn   & 89.0 & \textbf{91.1} & \textbf{87.3} & \textbf{88.8} & \textbf{89.1} & \textbf{92.0} & \textbf{93.1} & \textbf{88.6} & 89.8 & \textbf{90.9}\\
Reptile       & 88.1 & 87.9 & 86.8 & 87.5 & 87.6 & 90.6 & 91.7 & 87.3 & 86.2 & 89.0 \\ 
\bottomrule                       
\end{tabular}
\caption{{Average accuracy of 5 different seeds on the unseen target languages for Amazon.} $\Delta$ corresponds to the average accuracy across test languages.}
\label{tab:results_amazon_loo}
\end{table*}

\subsection{Baselines} \label{sec:baselines}
We introduce baselines trained in a standard supervised, non-episodic fashion. Again, we use  XLM-RoBERTa-base as the base-learner in all models.
\paragraph{Zero-shot} 
This baseline assumes sufficient training data for the task to be available in one language $l_{src}$ (English). The base-learner is trained in a non-episodic manner using mini-batch gradient descent with cross-entropy loss. Performance is monitored during training on a held-out validation set in $l_{src}$, the model with the lowest loss is selected, and then evaluated on the same task in the target languages. 

\paragraph{Non-episodic}
The second baseline aims to quantify the exact impact of learning a model through the meta-learning paradigm versus standard supervised learning. The model learns from exactly the same data as the meta-learning algorithms, but in a non-episodic manner: i.e., merging support and query sets in $l_{aux}$ (and $l_{src}$ when included) and training using mini-batch gradient descent with cross-entropy loss. During testing, the trained model is independently finetuned for 5 steps on the support set (one mini-batch) of each target language $l_{tgt}$. 

\subsection{Training setup and hyper-parameters}
We use the Ranger optimizer, an adapted version of Adam \cite{kingma2014adam} with improved stability at the beginning of training -- by accounting for the variance in adaptive learning rates \cite{liu2019variance} -- and improved robustness and convergence speed \cite{zhang2019lookahead,yong2020gradient}.
We use a batch size of 16 and a learning rate of 3e-5 to which we apply cosine annealing. 
For meta-training, we perform 100 epochs of 100 episodes and perform evaluation with 5 different seeds on the meta-validation set after each epoch. One epoch consists of 100 update steps where each update step consists of a batch of 4 episodes. Early-stopping with a patience of 3 epochs is performed to avoid overfitting. 
For the non-episodic baselines, we train for 10 epochs on the auxiliary languages while validating after each epoch. All models are created using the PyTorch library \cite{paszke2017automatic} and trained on a single 24Gb NVIDIA Titan RTX GPU.

We perform grid search on MLDoc in order to determine optimal hyperparameters for the MetaUpdate methods.
The hyper-parameters resulting in the lowest loss on $l_{dev} =$ Spanish are used in all experiments. 
The number of update steps in the inner-loop is 5; the (initial) learning rate of the inner-loop is 1e-5 for MAML and ProtoMAML and 5e-5 for Reptile; the factor by which the learning rate of the classification head is multiplied is 10 for MAML and ProtoMAML and 1 for Reptile; when applicable, the learning rate with which the inner-loop optimizer is updated is 6e-5. See Table \ref{tab:grid_search} for the considered grid. 

\begin{table*}[th]
\resizebox{\textwidth}{!}{
\centering
\resizebox{\textwidth}{!}{\begin{tabular}{llllll|l|llllll|l}
\toprule
\multicolumn{1}{c}{\multirow{2}{*}{$\mathbf{l_{src}}$ = {en}}} & \multicolumn{1}{c}{\multirow{2}{*}{\textbf{Method}}} & \multicolumn{5}{c}{\textbf{Amazon}} & \multicolumn{7}{c}{\textbf{MLDoc}}  \\
  & & \textbf{de}  & \textbf{fr} & \textbf{ja} & \textbf{zh} & $\Delta$ & \textbf{de}  & \textbf{fr} & \textbf{it} & \textbf{ja} & \textbf{ru} & \textbf{zh} & $\Delta$ \\
                       \midrule
\multirow{5}{*}{Excluded} 
& Non-episodic & 88.4 & 88.6 & 85.7 & 88.2 & 87.7 & 92.8 & 89.1 & 81.2 & 83.2 & \textbf{84.0} & 87.4 & 86.3 \\
& ProtoNet & 86.7 & 88.0 & 86.2 & 87.3 & 87.1 & 89.7 & 87.6 & 80.5 & 82.2 & 80.6 & 85.2 & 84.3\\
& foMAML & 88.3 & 87.5 & 84.6 & \textbf{89.1} & 86.3 & 94.1 & \textbf{89.7} & \textbf{81.5} & 84.2 & 77.6 & 87.5 & 85.8 \\
& foProtoMAMLn   & \textbf{88.9} & \textbf{89.5} & \textbf{86.5} & 89.0 & \textbf{88.5} & \textbf{94.8} & 89.5 & \textbf{81.5} & \textbf{84.8} & 81.0  &\textbf{88.7} & \textbf{86.6} \\
& Reptile       & 86.1 & 86.3 & 82.9 & 87.0 & 85.6 & 92.4 & 88.2 & 80.5 & 82.5 & 79.5 & 87.8 & 85.3 \\ 
\midrule
\multirow{5}{*}{Included} 
& Non-episodic & \textbf{91.0} & 91.0 & 87.3 & 89.4 & 89.8 & 94.9 & 92.1 & 84.7 & 84.8 & 83.7 & \textbf{91.4} & 88.6 \\
& ProtoNet & 90.3 & 91.3 & 87.5 & 88.7 & 89.5 & 95.5 & 91.7 & 83.4 & 85.1 & 82.8 & 88.3 & 87.8 \\
& foMAML & 90.1 & 90.7 & 87.2 & 89.5 & 89.4 & 95.1 & 92.5 & 83.1 & 84.9 & 84.3 & 90.6 & 88.4 \\
& foProtoMAMLn   & 90.7 & \textbf{91.5} & \textbf{88.0} & \textbf{90.4} & \textbf{90.2} & \textbf{96.0} & \textbf{93.6} & \textbf{85.0} & \textbf{85.7} & \textbf{84.8} & 90.8 & \textbf{89.3}  \\
& Reptile    &   90.0 & 89.5 & 86.5 & 87.6 & 88.4 & 94.4 & 93.1 & 83.8 & 85.2 & 83.6 & 90.4 & 88.4 \\ 
\bottomrule                       
\end{tabular}}}
\caption{{Average accuracy of 5 different seeds on the target languages in the joint-training setting for MLDoc and Amazon.} $\Delta$ corresponds to the average accuracy across test languages. }
\label{tab:results_joint}
\end{table*}

\section{Results} \label{sec:results}
\paragraph{Cross-lingual adaptation}
Tables \ref{tab:results_mldoc_half} and \ref{tab:results_amazon_loo} show the accuracy scores on the target languages on MLDoc and Amazon respectively. We start by noting the strong multilingual capabilities of XLM-RoBERTa as our base-learner:
Adding the full training datasets in three extra languages (i.e., comparing the zero-shot with the non-episodic baseline in the high-resource, `Included' setting) results in a mere 1.2\% points increase in accuracy on average for MLDoc and 0.6\% points for Amazon. 
Although the zero-shot\footnote{The zero-shot baseline is only applicable in the `Included' setting, as the English data is not available under `Excluded'.} and non-episodic baselines are strong, in the majority of cases, a meta-learning approach improves performance. This holds especially for our version of ProtoMAML (ProtoMAMLn), which achieves the highest average accuracy in all considered settings. 

The substantial improvements for Russian on MLDoc and Chinese on Amazon indicate that meta-learning is most advantageous when the considered task distribution is somewhat heterogeneous or, in other words, when \textit{domain drift} \cite{lai2019bridging} is present. For the Chinese data used for the sentiment polarity task, the presence of domain drift is obvious as the data is collected from a different website and concerns different products than the other languages. For Russian in the MLDoc dataset, it holds that the non-episodic baseline has the smallest gain in performance when adding English data ($l_{src}$) in the limited-resource setting (0.2\% absolute gain as opposed to 5.7\% on average for the remaining languages) and even a decrease of 2.4\% points when adding English data in the high-resource setting. Especially for these languages with domain drift, our version of ProtoMAML (foProtoMAMLn) outperforms the non-episodic baselines with a relatively large margin. For instance, in Table \ref{tab:results_mldoc_half} in the high-resource setting with English included during training, foProtoMAMLn improves over the non-episodic baseline with 9.1\% points whereas the average gain over the remaining languages is 0.9\% points. A similar trend can be seen in Table \ref{tab:results_amazon_loo} where, in the limited-resource setting, foProtoMAMLn outperforms the non-episodic baseline with 1.9\% points on Chinese, with comparatively smaller gains on average for the remaining languages.

\paragraph{Joint training}      
In this setting, we achieve a new state of the art on MLDoc for German, Italian, Japanese and Russian using our method, foProtoMAMLn (Table \ref{tab:results_joint}).\footnote{The zero-shot baselines are the same as in Tables \ref{tab:results_mldoc_half} and \ref{tab:results_amazon_loo}.} The previous state of the art for German and Russian is held by \citet{lai2019bridging} (95.73\% and 84.65\% respectively).
For Japanese and Italian, it 
is held by \citet{eisenschlos2019multifit} (80.55\% and 80.12\% respectively). The state of the art for French and Chinese is also held by \citet{lai2019bridging} (96.05\% and 93.32\% respectively).
On the Amazon dataset, foProtoMAMLn also outperforms all other methods on average. The state of the art is held by \shortcite{lai2019bridging} with 93.3\%, 94.2\% and 90.6\% for French, German and Chinese respectively and, although we do not outperform it, the differences are rather small -- between 0.2\% (Chinese) and 3.4\% points (German) -- even when grid search is based on MLDoc, while we use a much less computationally expensive approach. 

\begin{figure}[h]
    \centering
    \includegraphics[width=\linewidth]{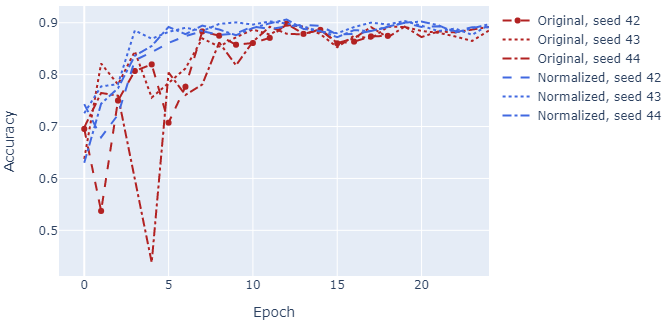}
    \caption{Validation accuracy for 3 seeds for original foProtoMAML and our new method, foProtoMAMLn.}
    \label{fig:protomaml_stability}
\end{figure}

Again, we use Russian in MLDoc to exemplify the difference between meta-learning and standard supervised learning. When comparing the difference in performance between excluding and including English meta-training episodes ($l_{src}$), opposite trends are noticeable: for standard supervised, non-episodic learning, performance drops slightly by 0.3\%, whereas all meta-learning algorithms gain between 2.2\% and 6.7\% in absolute accuracy. This confirms our earlier finding that meta-learning benefits from, and usefully exploits heterogeneity in data distributions; in contrast, this harms performance in the standard supervised-learning case.

\begin{table*}[t]
\centering
\resizebox{0.65\textwidth}{!}{
    \footnotesize
    
    \resizebox{\linewidth}{!}{
        \begin{tabular}{lllllll|l}
        \toprule
        \makecell{\textbf{Dataset}} & \textbf{de}  & \textbf{fr} & \textbf{it} & \textbf{ja} & \textbf{ru} & \textbf{zh} & Diff \\ 
                               \midrule
        Amazon & 90.4 & 90.9 & - & 87.3 &- & 88.3 & -1.7 \\
        MLDoc & 92.8 & 92.4 & 78.6 & 79.3 & 69.3 & 88.9 & -4.3 \\
        \bottomrule                       
        \end{tabular}
    }
}
\caption{{Average accuracy of 5 different seeds on unseen target languages using the original/unnormalized foProtoMAML model.} \textit{Diff} is the difference in average accuracy $\Delta$ across languages against foProtoMAMLn.}
\label{tab:unnormalized_protomaml}
\end{table*}

\begin{table*}[t]
\centering
\begin{tabular}{lllll|l|llll|l}
\toprule
 \multicolumn{1}{c}{\multirow{2}{*}{\textbf{Method}}} & \multicolumn{4}{c}{\textbf{Limited-resource setting}} & & \multicolumn{4}{c}{\textbf{High-resource setting}}  \\ & \textbf{de} & \textbf{fr} & \textbf{ja} & \textbf{zh} & Diff  & \textbf{de} & \textbf{fr} & \textbf{ja} & \textbf{zh} & Diff \\
                       \midrule
ProtoNet & 91.1 & 90.9 & 87.1 & 85.5 &  +0.75 & 91.3 & 91.1 & 87.4 & 88.7 &  +1.44\\
foMAML & 90.8 & 87.4 & 87.3 & 85.2 &  -0.75 & 91.7 & 91.2 & 87.2 & 88.1 &  -1.13\\
foProtoMAMLn   & 87.7 & 87.8 & 83.9 & 84.4 &  -3.1 & 90.8 & 89.8 & 86.2 & 82.3 &  -3.96 \\
Reptile       & 89.3 & 90.2 & 86.7 & 85.5 &  +0.35 & 90.0 & 89.3 & 87.1 & 85.7 &  -1.04 \\ 
\bottomrule                       
\end{tabular}
\caption{{Average accuracy of 5 different seeds on unseen target languages for Amazon when initializing from monolingual classifier in $l_{src}$.} \textit{Diff}: difference in average accuracy $\Delta$ across languages compared to initializing from the XLM-RoBERTa language model.}
\label{tab:results_amazon_loo_from_teacher}
\end{table*}

\section{Ablations}
\paragraph{foProtoMAMLn}
Figure \ref{fig:protomaml_stability} shows the development of the validation accuracy during training for 25 epochs for the original foProtoMAML and our model, foProtoMAMLn. By applying $L_2$ normalization to the prototypes, we obtain a more stable version of foProtoMAML which empirically converges faster. We furthermore re-run the high-resource experiments with English for both MLDoc and Amazon using the original foProtoMAML (Table \ref{tab:unnormalized_protomaml}) and find it performs 4.3\% and 1.7\% accuracy points worse on average, respectively, further demonstrating the effectiveness of our approach. 

\paragraph{Initializing from a monolingual classifier}
In our experiments, we often assume the presence of a source language (English). We now investigate (in the $l_{src}$ = {en} `Excluded' setting) whether it is beneficial to pre-train the base-learner in a standard supervised way on this source language and use the obtained checkpoint $\theta_{mono}$ as an initialization for meta-training (Table \ref{tab:results_amazon_loo_from_teacher}) rather than initializing from the transformer checkpoint. 

We observe that only ProtoNet consistently improves performance, whereas foProtoMAMLn suffers the most with a decrease of 3.1\% and 3.96\% in accuracy in the low- and high-resource setting respectively. 
We surmise this difference is attributable to two factors. Intuitively, the monolingual classifier aims to learn a transformation from the input space to the final feature space, from which the prototypes for ProtoNet and ProtoMAML are created, in which the learned classes are encoded in their own disjoint sub-spaces such that a linear combination of these features can be used to correctly classify instances. ProtoNet aims to learn a similar transformation, but uses a Nearest Neighbours approach to classify instances instead. ProtoMAML on the other hand benefits the most from prototypes which can be used to classify instances \textit{after} the inner-loop updates have been performed. This, in combination with the fact that the first-order approximation of ProtoMAML cannot differentiate through the creation of the prototypes, could explain the difference in performance gain with respect to ProtoNet.

\section{Conclusion}
We proposed a meta-learning framework for few-shot cross- and multilingual joint-learning for document classification tasks in different domains. We demonstrated that it leads to consistent gains over traditional supervised learning on a wide array of data availability and diversity settings, and showed that it thrives in settings with a heterogenous task distribution. We presented an effective adaptation to ProtoMAML and, among others, obtained a new state of the art on German, Italian, Japanese and Russian in the few-shot setting on MLDoc. 

\section{Acknowledgements}
This work was supported by Deloitte Risk Advisory B.V., the Netherlands.

\bibliography{eacl2021}
\bibliographystyle{acl_natbib}



\end{document}